\documentclass[runningheads]{llncs}

\usepackage{eccv}

\usepackage{eccvabbrv}

\usepackage{graphicx}
\usepackage{booktabs}
\usepackage{multirow}
\usepackage{pgf}

\usepackage[accsupp]{axessibility}  

\usepackage{hyperref}

\usepackage{orcidlink}

\newcommand{\cmark}{\ensuremath{\checkmark}}
\newcommand{\xmark}{\textendash}

\definecolor{visualred}{HTML}{C0392B}
\definecolor{audioblue}{HTML}{2874A6}
\newcommand{\vcue}[1]{\textcolor{visualred}{#1}}
\newcommand{\acue}[1]{\textcolor{audioblue}{#1}}

\begin{document}

\title{\texttt{TAKE\,85}: Testing Audiovisual filmmaKer's intEnt across 85 Hours of Film}

\titlerunning{TAKE 85}

\makeatletter
\newcommand{\samethanks}[1][\value{footnote}]{\footnotemark[#1]}
\makeatother

\author{Kaishuu Shinozaki-Conefrey\inst{1,2}\orcidlink{0009-0005-0943-5216}\thanks{Equal contribution.} \and
Olivier Pascaud\inst{1,3}\orcidlink{0009-0004-1803-4943}\samethanks \and
Robin Courant\inst{1}\orcidlink{0009-0002-5329-4009} \and
Xi Wang\inst{1}\orcidlink{0000-0001-6586-1926} \and
Dimitris Samaras\inst{4}\orcidlink{0000-0002-1373-0294} \and
Vicky Kalogeiton\inst{1}\orcidlink{0000-0002-7368-6993}}

\authorrunning{K.~Shinozaki-Conefrey et al.}

\institute{LIX, Ecole Polytechnique, IP Paris, Palaiseau, France \and
New York University, New York, NY, USA \and
Ecole nationale sup\'erieure Louis-Lumi\`ere, Saint-Denis, France \and
Stony Brook University, Stony Brook, NY, USA}
\maketitle

\vspace{-0.5cm}
\begin{figure}[h!]
    \centering
    \includegraphics[width=\textwidth]{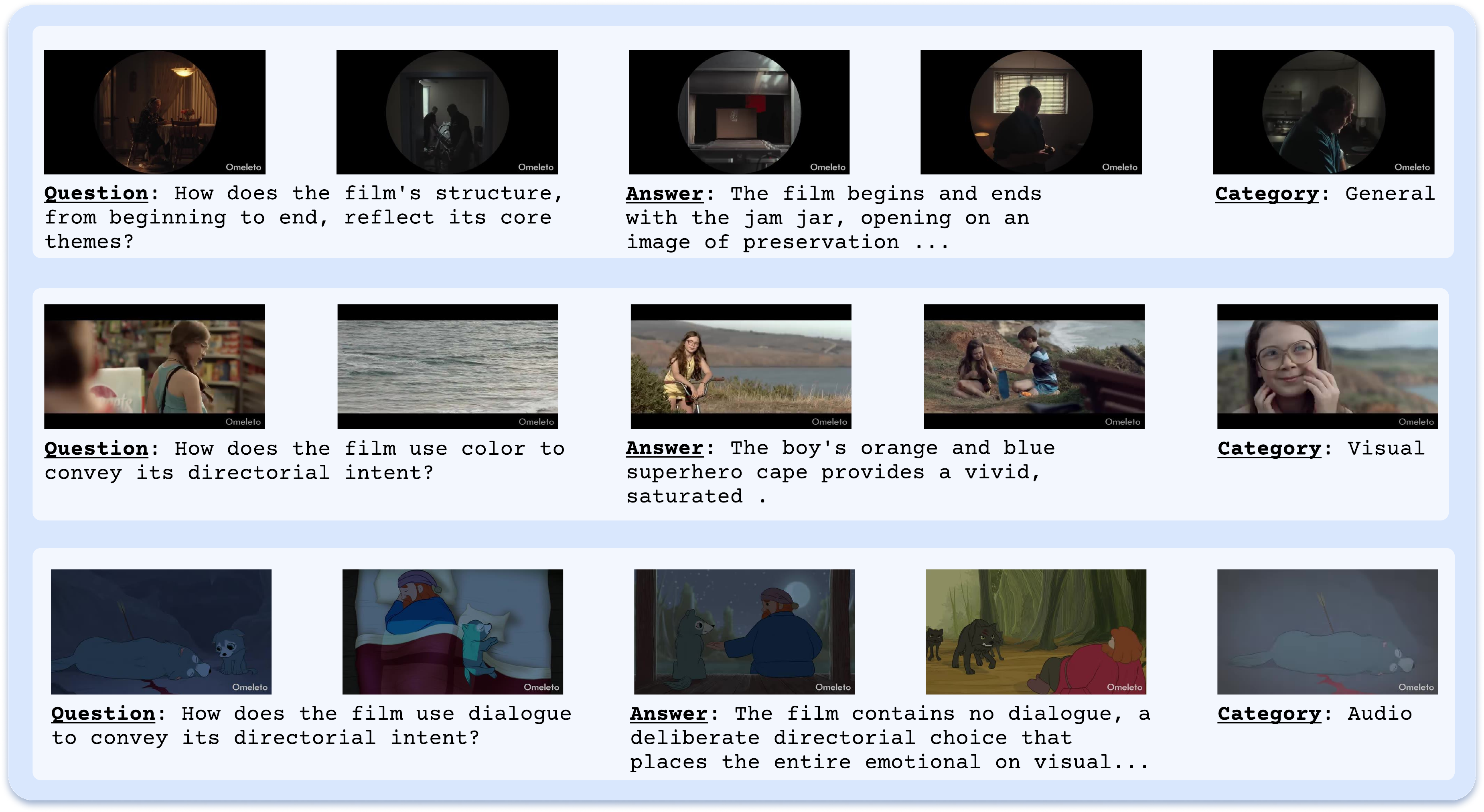}
    \caption{Example question--answer pairs from \texttt{TAKE\,85} spanning all three categories.}
    \label{fig:teaser}
\end{figure}
\vspace{-1cm}

\begin{abstract}
Films communicate through deliberate creative choices, including lighting, color, composition, editing, dialogue, music, and sound. Humans naturally interpret these signals as \emph{directorial intent}, yet current multimodal large language models (MLLMs) are evaluated almost exclusively on understanding \emph{what} happens rather than \emph{why} it is presented that way. We introduce \textbf{\texttt{TAKE\,85}}, the first benchmark for directorial-intent understanding, comprising 398 short films (85 hours) with expert-verified question--answer pairs spanning global and fine-grained visual and audio intent. Through controlled modality ablations, \texttt{TAKE\,85} enables systematic evaluation of multimodal reasoning. Experiments on state-of-the-art MLLMs reveal a substantial gap between perceptual recognition and intentional understanding: while models accurately describe events and narratives, they consistently fail to infer the communicative role of filmmaking decisions. Our results establish \textbf{directorial intent} as a previously overlooked dimension of
multimodal understanding: even the strongest model reaches only 58 out of 100, and our ablations show that no input modality is sufficient on its own. All code, Q\&As, and models are publicly available from \url{https://www.lix.polytechnique.fr/vista/projects/2026_take85_shinozaki/}.

\end{abstract}

\section{Introduction}
\label{sec:intro}
Films are carefully constructed acts of communication.
Every decision---lighting, color, staging, composition, editing, dialogue, music, sound design, pacing, and countless other creative choices---is deliberately orchestrated to influence how the audience experiences a story. These elements do far more than depict events. They establish mood, create suspense, reveal relationships, guide attention, and communicate ideas that are never explicitly stated. A single scene can evoke fear, intimacy, loneliness, or hope without changing a single line of dialogue, simply through the way it is crafted. Human viewers naturally decode this audiovisual language, often without conscious effort. 
We understand not only \emph{what} happens in a film, but also \emph{how} it is presented, shaping the viewing experience.
This hidden layer of meaning, conveyed through the filmmaker's creative decisions, is commonly referred to as \emph{directorial intent}.

Recent multimodal large language models (MLLMs) have made remarkable progress in video understanding~\cite{wang2024qwen2,bai2025qwen3,wang2025internvl3,zhang2025videollama,achiam2023gpt}. They can recognize objects and actions~\cite{krishna2017dense,yang2023vid2seq}, summarize long narratives~\cite{islam2024video,han2024autoad,zhang2024mm}, answer questions about events~\cite{xu2017video,yu2019activitynet}, and reason over hours of audiovisual content~\cite{wu2021towards,rawal2024cinepile,a2026cineaste}. Existing benchmarks reflect these advances by evaluating increasingly sophisticated perceptual capabilities~\cite{tang2025video,kumar2025videollm}, including temporal reasoning~\cite{li2022representation,zhou2024streaming}, plot comprehension~\cite{tapaswi2016movieqa,ghermi2024long}, visual question answering~\cite{lei2018tvqa,xu2017video}, and cinematographic technique recognition~\cite{tang2025vidcomposition,liu2026shotbench,wang2026cinetechbench}. As a consequence, current MLLMs appear to possess an impressive understanding of films~\cite{rawal2024cinepile,a2026cineaste,huang2020movienet}.

However, nearly all existing evaluations measure \emph{what} happens on screen rather than \emph{why} the filmmaker chose to present it that way. They reward recognition of events, characters, and dialogues, but largely ignore the communicative role of cinematographic and auditory decisions. Whether a scene is illuminated with cold blue lighting instead of warm tones, whether silence replaces music, whether characters are deliberately staged apart or close together, or whether rapid editing is used to increase tension, all fundamentally shape the viewer's interpretation. Yet these creative choices remain almost entirely absent from current multimodal benchmarks.

Understanding directorial intent presents a fundamentally different reasoning challenge from conventional video understanding. It cannot be solved by recognizing objects or describing isolated events, nor by considering visual or audio information independently. Instead, it requires reasoning over multiple modalities jointly to infer the communicative purpose behind the filmmaker's decisions. Consequently, directorial intent provides a natural and previously unexplored benchmark for measuring whether multimodal models truly integrate audiovisual information, rather than simply recognizing perceptual cues.

In this paper, we introduce \textbf{\texttt{TAKE\,85}}, the first benchmark dedicated to evaluating directorial intent understanding. \texttt{TAKE\,85} consists of 398 carefully curated short films (approximately 85 hours of content) together with expert annotations describing the intentions conveyed through both visual and auditory filmmaking decisions. Rather than evaluating films only at a global level, \texttt{TAKE\,85} decomposes directorial intent into fine-grained cinematic dimensions, including overall intent, visual intent, audio intent, lighting, color, composition, staging, dialogue, music, and sound design. To ensure high-quality supervision, annotations are produced in collaboration with film experts and grounded using multimodal evidence extracted from the visual stream, subtitles, music, and sound effects.

Beyond introducing a new benchmark, \texttt{TAKE\,85} enables systematic analysis of multimodal reasoning through controlled modality ablations. We evaluate state-of-the-art MLLMs under different combinations of visual, audio, and textual inputs, allowing us to quantify the contribution of each modality to directorial-intent understanding. Our experiments reveal a consistent and substantial gap between perceptual recognition and intentional reasoning. Although modern MLLMs successfully recognize objects, actions, and narrative events, they struggle to infer the communicative role of creative filmmaking decisions. In particular, they consistently fail on questions about sound design,
directorial technique, and how a film is composed and structured, and exhibit
limited ability to combine complementary information across modalities. These failures persist even when all available inputs are provided, suggesting that current multimodal models remain largely focused on describing \emph{what} they perceive rather than understanding \emph{why} a scene was constructed in that way.

Our findings expose a previously overlooked limitation of current multimodal foundation models. As MLLMs continue progressing from perception toward genuine understanding, evaluating directorial intent represents an essential next step. We hope \texttt{TAKE\,85} establishes this capability as a new benchmark for measuring higher-level audiovisual reasoning.

Our contributions are threefold:
\begin{enumerate}
    \item We introduce \textbf{\texttt{TAKE\,85}}, the first benchmark for evaluating directorial intent in films, comprising 398 short films with expert annotations spanning both global and fine-grained filmmaking decisions.
    \item We propose a multimodal evaluation framework that systematically studies directorial-intent understanding through controlled combinations of visual, audio, and textual inputs.
    \item We demonstrate that current state-of-the-art MLLMs consistently fail to understand directorial intent, revealing a significant gap between perceptual recognition and higher-level audiovisual reasoning.
\end{enumerate}

\section{Related work}
\label{sec:rel_work}
\paragraph{Video understanding.}
Understanding video content is a long-standing problem in computer
vision~\cite{tang2025video,kumar2025videollm}.
A first line of work, \emph{dense video captioning}, jointly localizes and describes the events in a video: introduced with the ActivityNet Captions benchmark~\cite{krishna2017dense}, later made end-to-end~\cite{zhou2018end,wang2018bidirectional}, then scaled to pretrained visual language models~\cite{yang2023vid2seq}, hour-long recursive summarization~\cite{islam2024video}, and streaming settings~\cite{zhou2024streaming}. These methods describe factual content (\textit{what, when, where, and who}), but completely ignore the narrative aspect.
Audio Description (AD) adds a narrative layer, generating spoken descriptions of salient visuals for visually impaired audiences.
The AutoAD line conditions a language model on surrounding context and prior descriptions, adding character naming and speak-timing before moving back to raw pixels~\cite{han2023autoad,han2023autoadii,han2024autoad}; follow-ups pursue zero-shot~\cite{xie2024autoad}, in-context~\cite{zhang2024mm}, distinctive~\cite{fang2025distinctad}, unified~\cite{wang2025contextual}, and film-grammar-aware~\cite{xie2025shot} generation.
Still, AD remains about what is depicted (\textit{who, what, where}), not why the filmmaker chose to depict it that way.
Finally, more recently, general-purpose multimodal large language models (MLLMs)~\cite{wang2024qwen2,bai2025qwen3,wang2025internvl3,zhang2025videollama,achiam2023gpt} handle video within a broad instruction-following interface with strong perceptual and descriptive abilities. Yet they remain perceptual and only describe content.
Across all three lines of work, one aspect remains unexplored: the cinematic intent behind a scene, i.e., why the filmmaker staged it that way. In this work, we take a first step in this direction and evaluate how well MLLMs understand cinematic intent.

\paragraph{Datasets and Benchmarks.}
A first family of benchmarks probes factual and plot-level comprehension through video question answering.
Early datasets target open-domain clips: MSVD-QA~\cite{xu2017video} and MSRVTT-QA~\cite{xu2017video} generate QA pairs automatically to test appearance and motion, while ActivityNet-QA~\cite{yu2019activitynet} scales this to long web videos with human-written questions.
TVQA~\cite{lei2018tvqa} grounds compositional questions in TV shows that require jointly analyzing frames and subtitles, and Causal-VidQA moves beyond recognition to explanatory, predictive, and counterfactual reasoning~\cite{li2022representation}.
At the movie scale, MovieQA tests story comprehension from video, subtitles, and scripts~\cite{tapaswi2016movieqa}, LVU introduces long-form tasks such as relationship, speaking style, and genre prediction~\cite{wu2021towards}. More recent efforts push both scale and difficulty: CinePile~\cite{rawal2024cinepile} builds 300K multiple-choice questions demanding genuine long-range comprehension, SF20K~\cite{ghermi2024long} draws story-level questions from 20K short films, and Cin\'easte~\cite{a2026cineaste} targets fine-grained contextual reasoning across full-length movies. Yet these remain anchored in \emph{what happens}, i.e., the plot and its causal structure.
A second family instead probes cinematic \emph{craft} itself, i.e., how a scene is filmed.
VidComposition~\cite{tang2025vidcomposition} asks whether MLLMs can analyze the composition and editing of compiled videos, and CineCap~\cite{mao2026cinecap} targets structured captioning of camera movement, shot size, and depth of field.
Dedicated benchmarks probe expert cinematographic grammar: ShotBench~\cite{liu2026shotbench} with expert-annotated QA over eight shot-level dimensions from acclaimed films, CineTechBench~\cite{wang2026cinetechbench} across shot scale, angle, composition, movement, lighting, color, and focal length, and CameraBench~\cite{lin2026towards} with a cinematographer-designed taxonomy of camera-motion primitives. These reveal that even the strongest MLLMs struggle to identify cinematographic technique.

\paragraph{Cinematography understanding.}
A complementary line of work analyzes films through their formal, cinematographic properties rather than their plot.
A large body of research classifies individual shots by scale, angle, and camera movement, from early hand-crafted descriptors~\cite{canini2013classifying} of composition, color, and motion to learning-based approaches~\cite{rao2020unified,jiang2021jointly,vacchetti2022cinematographic,savardi2023recognition,li2023toward,lu2024exploring}.
Rao~et~al.~\cite{rao2020unified} introduce the MovieShots dataset and a subject-guided network that jointly recognizes shot scale and movement, later extended by jointly learning shot attributes for boundary detection~\cite{jiang2021jointly}, ensemble classifiers over finer field-size categories~\cite{vacchetti2022cinematographic}, camera angle and level recognition from single frames~\cite{savardi2023recognition}, unified shot-attribute analysis~\cite{li2023toward}, and explainable, SAM-guided shot-type classification~\cite{lu2024exploring}.
Beyond individual shots, a few works move toward the intent behind a film. MovieNet~\cite{huang2020movienet} is a first step in this direction, a holistic dataset pairing footage with cast, scripts, and stylistic annotations. Building on such resources, trailers provide weak supervision for movie understanding~\cite{huang2018trailers} and high-level cinematographic features characterize directorial style~\cite{courant2021high}.
Humor detection~\cite{liu2022funnynet,liu2024funnynet,barriere2025standup4ai,hanania2026mtllfm} is an especially interesting proxy for implicit intent since it might not be stated directly and relies on the interplay of multimodal cues, so recognizing it requires reading beyond the literal content.
All of these, however, \emph{detect} stylistic or affective attributes rather than \emph{interpret} them: they label how a film is shot or whether a moment is funny, but never the intent behind those choices.
In contrast, we introduce a benchmark that directly probes this intent, evaluating whether models can reason about \emph{why} a scene is staged and shot the way it is.

\section{\texttt{TAKE\,85}: A directorial intent dataset}
\label{sec:dataset}
\begin{table}[t]
\centering
\scriptsize
\caption{\textbf{\texttt{TAKE\,85} statistics.} Questions are labelled general, visual or audio.}
\label{tab:stats}
\begin{tabular}{@{}ccccc@{}}
\toprule
 & & \multicolumn{2}{c}{Questions per film} & \\
\cmidrule(lr){3-4}
Number of films & Total footage (hours) & Simple & Specific & Total Q\&A pairs \\
\midrule
398 & 85.94 & 3 & 9 & 4{,}776 \\
\bottomrule
\end{tabular}
\end{table}

\texttt{TAKE\,85} comprises 398 short films drawn from SF20K~\cite{ghermi2024long}, totaling approximately
85 hours of footage (details in Table~\ref{tab:stats}). 
Each film comes with two summaries: a summary of the plot and a short intent metadata summary from the film's director, referred to as \emph{hand-written intent metadata}.

\begin{figure}[htp!]
    \centering
    \includegraphics[width=\textwidth]{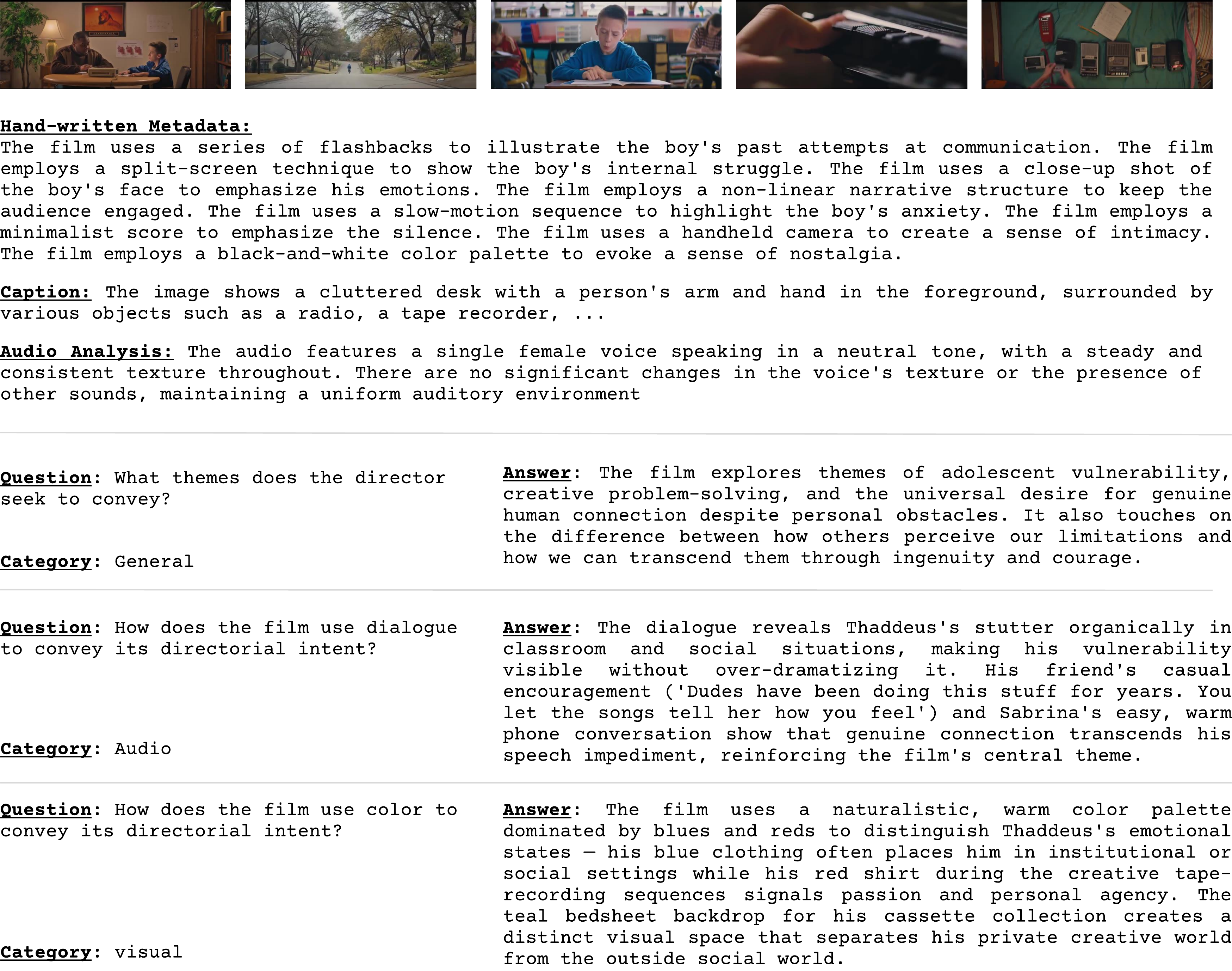}
    \caption{\textbf{A data sample of \texttt{TAKE\,85}.} Sampled keyframes, the inputs used to synthesize the ground truth, one visual caption, one audio-analysis and three Q\&A pairs.}
    \label{fig:demo1}
\end{figure}

For each film, \texttt{TAKE\,85} provides 12 directorial-intent question--answer pairs, yielding $4{,}776$ Q\&A pairs. 
The question set is fixed per movie, from global to fine-grained elements, spanning audio, text, and visual modalities.  
The answers come from the director's hand-written metadata, enriched with LLMs and manually verified by human cinema experts. 
The annotation process is described in Section~\ref{subsec:qa}. Figure~\ref{fig:demo1} shows one complete sample.

\subsubsection{Movie selection.}
\label{sub:selection}
We select films from SF20K's 20,143-film pool through multi-stage metadata filtering. We first restrict candidates to films sourced from Omeleto YouTube channel~\cite{omeleto_youtube}, a curated short-film distributor that provides \emph{hand-written intent metadata} for each of their films, yielding 1,649 candidates. 
An LLM then judges each description's directorial-intent strength (1: weak,
2: partial, 3: strong) and, separately, whether it addresses visual intent,
audio intent, both, or neither. Of the 1,649 films, 981 (59.5\%) score strong,
256 (15.5\%) partial and 412 (25.0\%) weak; 479 address both, 1,006 visual only,
30 audio only and 134 neither. Keeping those that address both and score at
least 2 yields the final set of 398 films, resulting in approximately 85 hours of footage.

\begin{figure}[ht]
    \centering
    \includegraphics[width=\textwidth]{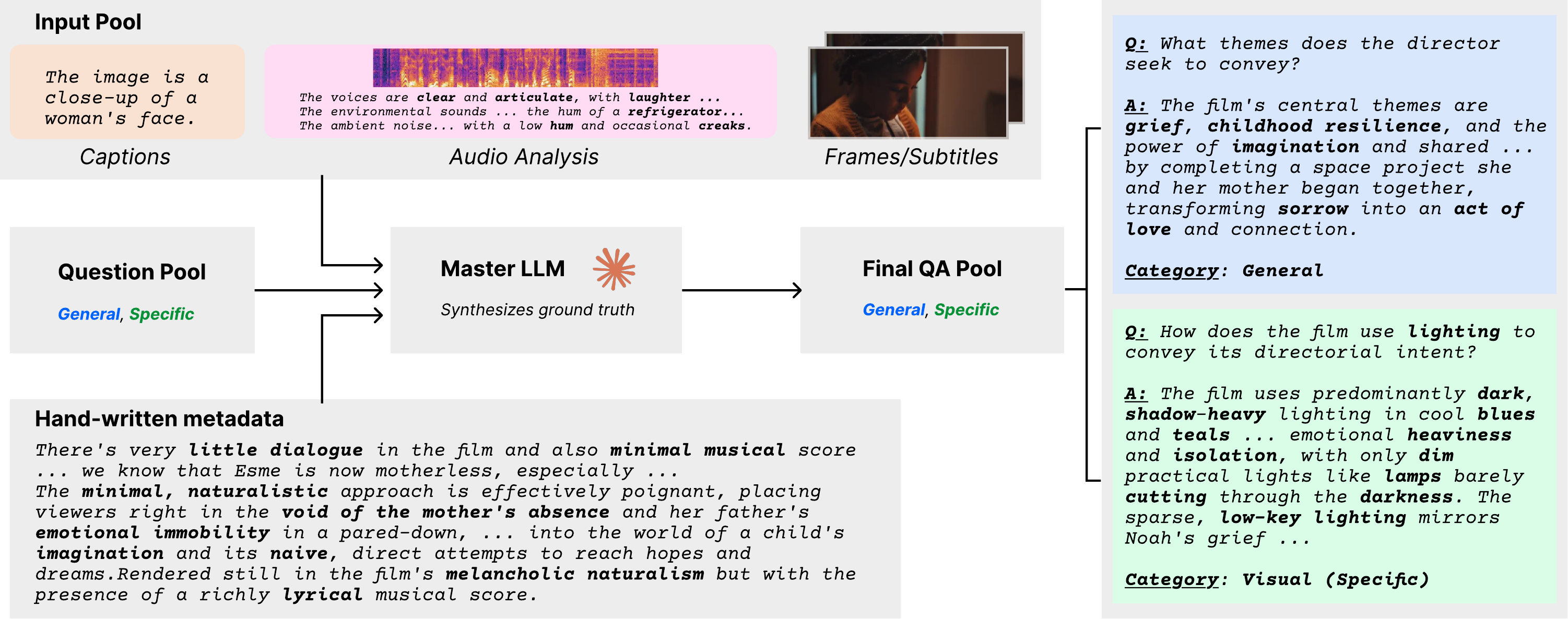}
    \caption{\textbf{Constructing \texttt{TAKE\,85}.} Visual captions, audio analysis, subtitles, and keyframes are extracted automatically from each film and combined with the hand-written intent metadata describing its intended themes and techniques. A master LLM answers the fixed pool of 12 intent questions from this evidence, producing the final QA pool. The creator metadata is used only here and is never shown to evaluated models.}
    \label{fig:pipeline}
\end{figure}

\subsection{Dataset pre-processing}
\label{sub:preprocessing}
For each film, we extract visual captions, audio, subtitles, and keyframes. Below, we detail the extraction of each element.

\paragraph{Visual captions.}
Visual captions and shot boundaries are taken from the original SF20K~\cite{ghermi2024long} metadata, with captions generated using Llama-3.2-Vision-11B~\cite{meta2024llama} and shot boundaries detected using PySceneDetect~\footnote{\url{https://github.com/Breakthrough/PySceneDetect}}.
Films average 142.9 shots (median 136.5, range 1–405). Shot boundary metadata is unavailable for 44 of the 398 films (11.1\%), for which keyframes are instead sampled at uniformly random timestamps rather than shot midpoints. Each caption describes the visual content of a representative frame within its shot (e.g. subjects, setting, composition, on-screen text), without reference to audio or narrative context. Captions for a film are concatenated in shot order, prefixed with a shot index (e.g. \texttt{[Shot 12]}), before being provided as model input.

\paragraph{Audio analysis.}
Audio analysis is generated by AudioFlamingo3~\cite{goel2025audio}, which independently describes a film's musical affect and sound-effect content in three equally-distributed temporal chunks of 30 seconds each. 
Prior to analysis, each film's audio track is separated into music and sound-effect stems using Meta's SAM-Audio~\cite{shi2025sam}, ensuring AudioFlamingo3 receives a clean, isolated signal for each rather than a mixed audio track. For music, AudioFlamingo3 produces both a description of the primary sonic elements (e.g. instrumentation, tempo, texture) and a separate description of the emotional affect the music conveys (e.g. ``melancholy and tense''). Sound-effect content is analyzed as a separate pass over its isolated stem, producing a numbered list of textual descriptions of ambient and diegetic sound events (e.g. footsteps, environmental noise, object interactions) independent of the music analysis. Both music and sound effects are grouped under the single ``audio analysis'' input channel.

\paragraph{Subtitles.}
Subtitles are transcribed using OpenAI's Whisper (large-v3-turbo)~\cite{radford2023robust}, applied directly to each film's audio track. 
Because automatic transcription
introduces errors, particularly on films with heavy background music, poor audio quality, or overlapping dialogue, we filter low-confidence segments before use: a subtitle segment is discarded if its Whisper no-speech probability exceeds 0.5 or its average log-probability falls below $-1.0$. Retained segments are timestamped to the nearest second and concatenated in chronological order.

\paragraph{Keyframes.}
We sample 20 keyframes per film. Where shot boundary information is available, one frame is sampled from the temporal midpoint of each of 20 evenly-spaced shots across the film, rather than from shot boundaries, to avoid capturing transitional or low-information frames. For the small number of films lacking shot boundary metadata, 20 timestamps are instead sampled uniformly at random across the film's duration.

\subsection{Q\&A construction}
\label{subsec:qa}

\paragraph{Questions.}
The question set is fixed rather than generated per film, so that every model answers the same questions about every film and scores remain directly comparable
across the benchmark. 
\begin{itemize}
    \item \textbf{Simple.} 3x questions about the overall, visual, and audio directorial intent;
    \item \textbf{Specific.}  9x questions about directorial devices, grouped into 3 categories:

\begin{enumerate}
\item \textbf{General}: techniques used, themes, and narrative structure.
\item \textbf{Visual}: lighting, color, and shot composition.
\item \textbf{Audio}: music, sound design, and dialogue.
\end{enumerate}
\end{itemize}

\noindent
This yields $398 \times 12 = 4{,}776$ questions. The category labels (general, visual, audio) are orthogonal to the global/fine-grained split, so scores
can be broken down along either axis independently.

\paragraph{Answers.}
Ground-truth construction begins and ends with human input, with an LLM
bridging the two. We start from hand-written intent metadata containing each film's intended themes and directorial techniques, provided by the film's creator.
An LLM enriches and complements it with detail drawn from the film's specific visual, audio, and textual evidence, grounding each claim in the modality that actually supports it. We close the loop with human
expert annotators, who review the films and the LLM-synthesized answers before they are used as ground truth.

Ground-truth answers are synthesized using Claude Sonnet 4.6~\cite{claude_sonnet} (see Figure~\ref{fig:pipeline})
from the complete context available for a film in a single call: (i) visual captions, (ii) audio analysis, (iii) subtitles,
(iv) 20 sampled keyframes (provided as images), and (v) the film's
\emph{hand-written intent metadata}.

All 12 questions for a film are answered in one call, with the model returning a
JSON object that maps each question number to its answer. Drawing on several
independently-derived sources rather than on a single modality is meant to keep
the answers specific to the film and checkable against its evidence, instead of
generic statements that would fit many films. A film's combined input ranges from
roughly 300 to 14,200 tokens (median $\approx$4,900, 99th percentile
$\approx$12,200), comfortably within the context window used for generation.

\paragraph{Human validation.}
LLM-synthesized answers are reviewed by \emph{human cinema experts} before being finalized as ground truth. For each film, the annotators watch the film in full, then judge each of the film's 12 question--answer pairs independently, marking each as correct or incorrect. A pair is marked incorrect if any part of the stated answer is factually wrong or unsupported by the film, prioritizing precision over partial credit. 

We validated 87 films in this way, covering 1,044 question--answer pairs. Only three pairs (0.3\%) were marked as incorrect; in each case, the annotators' comment identified the specific factual error (e.g.\ a claim about an event that does not occur in the film). The remaining 99.7\% were confirmed correct, with a small number carrying minor stylistic comments (e.g.\ imprecise word choice) that did not affect the correctness judgement.

\section{Evaluation protocol}
\label{sec:eval-protocol}
\paragraph{Task setup.}
Every model is asked the same 12 directorial-intent questions per film (Section~\ref{subsec:qa}), and is given only the input channels derived from the film itself: visual captions, audio analysis, subtitles, and the 20 sampled keyframes. The \emph{hand-written intent metadata} used to synthesize the ground truth (Section~\ref{sec:dataset}) is never shown to an evaluated model, so no model has access to a statement of the director's intent and must instead infer it from the film's audiovisual evidence. Two properties keep the comparison fair. First, channels are supplied separately, each under an explicit label, rather than fused into a single representation, so that any non-empty subset can be withheld without changing the format of the remaining input. Second, each question is answered independently: a model never sees its own answers to the other questions about the same film, which prevents a strong answer to the overall-intent question from propagating into the nine fine-grained ones. Section~\ref{sec:main-results} reports every model with all four channels available; Section~\ref{sec:modality-ablation} varies the subset, including the text-only regime in which the keyframes are withheld and a model must rely on the captions and audio analysis as a textual proxy for the film's imagery and soundtrack.

\paragraph{Metrics.}

Answers are free-form paragraphs rather than short spans, so we score them with an LLM judge and report two metrics. 
\begin{itemize}
    \item \textbf{Single-pass graded} asks the judge to rate, on a 1--5 scale, how well the prediction conveys the same substantive information as the reference, independently of wording, structure, and level of detail. In the final results, scores are rescaled to 0--100.

    \item \textbf{Checklist} decomposes each reference answer once into a set of atomic, independently verifiable claims; the decomposition is cached and reused across all models, so every model is graded against an identical rubric. The judge then decides, for each claim, whether the prediction supports it, and the score is the fraction of supported claims rescaled to 0--100. 
\end{itemize}

Single-pass graded uses a single holistic judgement per prediction, whereas the checklist replaces it with several narrower ones. Scores are broken down per question, following the structure of Section~\ref{sec:dataset}: the three simple questions (overall, visual, audio intent) and the nine specific ones grouped into general (techniques, themes, structure), visual (lighting, color, composition), and audio (music, sound design, dialogue).

\section{\texttt{TAKE\,85} benchmark}
\label{sec:results}
This section evaluates how well current multimodal models recover directorial
intent on \texttt{TAKE\,85}: overall performance with all input channels
available (Section~\ref{sec:main-results}), the contribution of each channel in
isolation (Section~\ref{sec:modality-ablation}), and which questions every model
answers well or fails (Section~\ref{sec:quali}).

\paragraph{Baselines.}
We evaluate ten multimodal large language models. Three are open-weight families spanning roughly an order of magnitude in parameter count, InternVL3.5~\cite{wang2025internvl3} (4B, 8B, 30B-A3B), Qwen3.5~\cite{qwen3.5} (4B, 9B, 27B), and Gemma~4~\cite{team2026gemma} (E4B, 12B, 26B-A4B), which we group into matched \emph{tiny}, \emph{small}, and \emph{large} bins; Gemini~3.5 Flash~\cite{google_gemini_3_5_flash} serves as a proprietary reference. Evaluating a scaling ladder within each family, and matched capacities across families, lets us separate the effect of model scale from that of training recipe. Because every model accepts all four channels, the same set of models supports both the full-context results of Section~\ref{sec:main-results} and the modality ablation of Section~\ref{sec:modality-ablation}, so \emph{differences between input conditions are never confounded with differences in architecture}.

\paragraph{Inference details.}
Open-weight models are served locally with vLLM via an OpenAI-compatible API, and Gemini~3.5 Flash through its provider API. All models are decoded greedily (temperature 0) and share one prompt template, so only the available channels vary between runs. The context window is sized to the longest input observed across all channel combinations (Section~\ref{sec:dataset}), so no film is truncated. The same judge model and prompts are used throughout.

\subsection{Directorial-intent understanding with full context}
\label{sec:main-results}

\begin{table*}[t]
\centering
\caption{\textbf{Main results on \texttt{TAKE\,85}} with all modalities provided
(\vcue{visual captions}, \vcue{sampled keyframes}, \acue{subtitles},
\acue{audio analysis}). Throughout, \vcue{visual} is red and \acue{audio} is blue.
\emph{Simple} reports the three overall directorial-intent questions
(G: overall, \vcue{V}: visual, \acue{A}: audio); \emph{Specific} the nine
fine-grained ones: general (G1--G3: techniques, themes, structure),
\vcue{visual} (V1--V3: lighting, color, composition), \acue{audio}
(A1--A3: music, sound design, dialogue). Avg. columns give each subgroup's mean
and, last, the mean over all nine.
The winning score per question category across each size category (except for \emph{Proprietary}) is put in bold.
\textbf{Single-pass graded score (0--100).}}
\label{tab:sota-judgement}
\scriptsize
\setlength{\tabcolsep}{2pt}
\newcommand{\modelindent}{\hspace{1em}}
\resizebox{\textwidth}{!}{%
\begin{tabular}{l @{\hspace{2.5em}} cccc @{\hspace{2.5em}} cccc @{\hspace{1.5em}} cccc @{\hspace{1.5em}} cccc @{\hspace{2.5em}} c}
\toprule
\multicolumn{1}{c@{\hspace{2.5em}}}{\multirow{2}{*}{\textbf{Model}}}
      & \multicolumn{4}{c@{\hspace{2.5em}}}{\textbf{Simple}} & \multicolumn{13}{c}{\textbf{Specific}} \\
\cmidrule(l{0pt}r{2.5em}){2-5} \cmidrule(l{0pt}r{0pt}){6-18}
      & G & \vcue{V} & \acue{A} & Avg.
      & G1 & G2 & G3 & Avg.
      & \vcue{V1} & \vcue{V2} & \vcue{V3} & \vcue{Avg.}
      & \acue{A1} & \acue{A2} & \acue{A3} & \acue{Avg.} & Avg. \\
\midrule
\multicolumn{18}{l}{\textit{Tiny}} \\
\modelindent InternVL3.5-4B   &  16.5 &  21.2 &  32.6 &  23.4 &  22.3 &  33.2 &  17.7 &  24.4 &  26.2 &  21.5 &  23.9 &  23.9 &  46.0 &  16.6 &  16.7 &  26.4 &  24.9 \\
\modelindent \textbf{Qwen3.5-4B}       &  \textbf{38.6} &  35.3 &  \textbf{44.2} &  \textbf{39.3} &  \textbf{31.8} &  \textbf{42.3} &  \textbf{36.9} &  \textbf{37.0} &  \textbf{42.6} &  \textbf{38.9} &  \textbf{34.0} &  \textbf{38.5} &  \textbf{47.1} &  \textbf{18.6} &  \textbf{31.0} &  \textbf{32.2} &  \textbf{35.9} \\
\modelindent Gemma~4 E4B      &  22.5 &  \textbf{36.6} &  42.3 &  33.8 &  30.2 &  37.2 &  22.6 &  30.0 &  32.9 &  32.2 &  21.5 &  28.9 &  43.2 &  12.9 &  29.8 &  28.6 &  29.2 \\
\addlinespace
\multicolumn{18}{l}{\textit{Small}} \\
\modelindent InternVL3.5-8B   &  29.4 &  32.5 &  40.7 &  34.2 &  24.9 &  38.3 &  16.3 &  26.5 &  30.7 &  28.7 &  37.1 &  32.2 &  47.7 &  21.7 &  24.5 &  31.3 &  30.0 \\
\modelindent \textbf{Qwen3.5-9B}       &  \textbf{45.8} &  \textbf{40.4} &  \textbf{47.4} &  \textbf{44.5} &  \textbf{30.5} &  \textbf{47.6} &  \textbf{35.4} &  \textbf{37.9} &  \textbf{40.7} &  \textbf{38.6} &  \textbf{41.6} &  \textbf{40.3} &  \textbf{52.6} &  \textbf{22.2} &  28.6 &  \textbf{34.5} &  \textbf{37.5} \\
\modelindent Gemma~4 12B      &  37.2 &  34.5 &  39.9 &  37.2 &  30.3 &  43.3 &  33.5 &  35.7 &  26.8 &  35.3 &  19.1 &  27.1 &  45.6 &  16.2 &  \textbf{35.9} &  32.6 &  31.8 \\
\addlinespace
\multicolumn{18}{l}{\textit{Large}} \\
\modelindent InternVL3.5-30B  &  37.4 &  34.0 &  41.7 &  37.7 &  25.1 &  41.0 &  27.9 &  31.3 &  40.7 &  27.3 &  44.8 &  37.6 &  54.6 &  22.4 &  27.3 &  34.8 &  34.6 \\
\modelindent \textbf{Qwen3.5-27B}      &  48.6 &  \textbf{45.6} &  \textbf{44.8} &  \textbf{46.4} &  \textbf{34.4} &  \textbf{52.6} &  \textbf{45.5} &  \textbf{44.2} &  \textbf{60.9} &  \textbf{43.2} &  \textbf{56.2} &  \textbf{53.4} &  \textbf{56.3} &  \textbf{28.1} &  36.2 &  \textbf{40.2} &  \textbf{45.9} \\
\modelindent Gemma~4 26B      &  \textbf{49.0} &  42.0 &  43.2 &  44.7 &  34.0 &  52.0 &  42.3 &  42.8 &  38.3 &  40.7 &  35.1 &  38.0 &  49.4 &  18.7 &  \textbf{44.6} &  37.6 &  39.5 \\
\addlinespace
\multicolumn{18}{l}{\textit{Proprietary}} \\
\modelindent Gemini~3.5 Flash &  70.3 &  54.8 &  60.9 &  62.0 &  47.8 &  70.1 &  65.8 &  61.2 &  62.6 &  51.2 &  58.2 &  57.4 &  66.1 &  37.1 &  64.0 &  55.7 &  58.1 \\
\bottomrule
\end{tabular}}
\end{table*}

\begin{table*}[t]
\centering
\caption{\textbf{Main results on \texttt{TAKE\,85}} with all modalities provided
(\vcue{visual captions}, \vcue{sampled keyframes}, \acue{subtitles},
\acue{audio analysis}). \emph{Simple} reports the three overall
directorial-intent questions (G: overall, \vcue{V}: visual, \acue{A}: audio);
\emph{Specific} the nine fine-grained ones --- general (G1--G3: techniques,
themes, structure), \vcue{visual} (V1--V3: lighting, color, composition),
\acue{audio} (A1--A3: music, sound design, dialogue). Avg. columns give each
subgroup's mean and, last, the mean over all nine.
The winning score per question category across each size category (except for \emph{Proprietary}) is put in bold.
\textbf{Checklist scores (0--100).}}
\label{tab:sota-checklist}
\scriptsize
\setlength{\tabcolsep}{2pt}
\newcommand{\modelindent}{\hspace{1em}}
\resizebox{\textwidth}{!}{%
\begin{tabular}{l @{\hspace{2.5em}} cccc @{\hspace{2.5em}} cccc @{\hspace{1.5em}} cccc @{\hspace{1.5em}} cccc @{\hspace{2.5em}} c}
\toprule
\multicolumn{1}{c@{\hspace{2.5em}}}{\multirow{2}{*}{\textbf{Model}}}
      & \multicolumn{4}{c@{\hspace{2.5em}}}{\textbf{Simple}} & \multicolumn{13}{c}{\textbf{Specific}} \\
\cmidrule(l{0pt}r{2.5em}){2-5} \cmidrule(l{0pt}r{0pt}){6-18}
      & G & \vcue{V} & \acue{A} & Avg.
      & G1 & G2 & G3 & Avg.
      & \vcue{V1} & \vcue{V2} & \vcue{V3} & \vcue{Avg.}
      & \acue{A1} & \acue{A2} & \acue{A3} & \acue{Avg.} & Avg. \\
\midrule
\multicolumn{18}{l}{\textit{Tiny}} \\
\modelindent InternVL3.5-4B   &   0.9 &   0.6 &   5.1 &   2.2 &   1.4 &   7.5 &   1.8 &   3.6 &   2.4 &   2.0 &   2.8 &   2.4 &  \textbf{13.1} &  \textbf{1.8} &   2.0 &   5.6 &   3.9 \\
\modelindent \textbf{Qwen3.5-4B}       &  \textbf{7.1} &  \textbf{3.1} &   7.2 &  \textbf{5.8} &  \textbf{2.9} &  \textbf{13.7} &  \textbf{4.9} &  \textbf{7.2} &  \textbf{9.8} &  \textbf{6.2} &  \textbf{6.7} &  \textbf{7.6} &  12.1 &   1.5 &  \textbf{6.3} &  \textbf{6.7} &  \textbf{7.1} \\
\modelindent Gemma~4 E4B      &   2.1 &   2.1 &  \textbf{8.8} &   4.4 &   2.0 &   7.8 &   2.5 &   4.1 &   4.5 &   3.6 &   3.3 &   3.8 &  11.2 &   1.0 &   2.8 &   5.0 &   4.3 \\
\addlinespace
\multicolumn{18}{l}{\textit{Small}} \\
\modelindent InternVL3.5-8B   &   3.2 &   1.6 &   8.9 &   4.6 &   1.1 &   8.1 &   0.9 &   3.4 &   3.3 &   2.5 &   5.0 &   3.6 &  13.1 &   2.2 &   3.3 &   6.2 &   4.4 \\
\modelindent \textbf{Qwen3.5-9B}       &  \textbf{9.2} &  \textbf{3.9} &  \textbf{10.5} &  \textbf{7.9} &  \textbf{2.6} &  \textbf{16.3} &  \textbf{5.9} &  \textbf{8.3} &  \textbf{8.3} &   5.1 &  \textbf{8.9} &  \textbf{7.4} &  \textbf{15.6} &  \textbf{2.3} &   5.2 &  \textbf{7.7} &  \textbf{7.8} \\
\modelindent Gemma~4 12B      &   5.5 &   2.6 &   9.2 &   5.7 &   2.4 &  13.3 &   3.4 &   6.3 &   4.8 &  \textbf{5.2} &   3.7 &   4.6 &  13.6 &   2.1 &  \textbf{6.4} &   7.4 &   6.1 \\
\addlinespace
\multicolumn{18}{l}{\textit{Large}} \\
\modelindent InternVL3.5-30B  &   5.7 &   2.1 &   9.4 &   5.7 &   1.7 &  10.3 &   3.2 &   5.0 &   4.5 &   2.3 &   4.5 &   3.8 &  15.6 &   2.4 &   4.9 &   7.6 &   5.5 \\
\modelindent \textbf{Qwen3.5-27B}      &  10.2 &  \textbf{5.4} &   9.9 &   8.5 &   3.2 &  \textbf{18.5} &  \textbf{8.2} &  \textbf{10.0} &  \textbf{18.3} &  \textbf{7.4} &  \textbf{15.7} &  \textbf{13.8} &  \textbf{17.8} &  \textbf{4.0} &   7.2 &  \textbf{9.7} &  \textbf{11.1} \\
\modelindent Gemma~4 26B      &  \textbf{10.7} &   4.0 &  \textbf{11.0} &  \textbf{8.6} &  \textbf{3.4} &  17.5 &   5.9 &   8.9 &   5.7 &   6.8 &   5.8 &   6.1 &  14.5 &   1.8 &  \textbf{9.2} &   8.5 &   7.8 \\
\addlinespace
\multicolumn{18}{l}{\textit{Proprietary}} \\
\modelindent Gemini~3.5 Flash &  23.9 &   9.6 &  19.6 &  17.7 &   8.4 &  33.9 &  18.9 &  20.4 &  19.4 &  13.8 &  17.8 &  17.0 &  25.3 &   6.4 &  19.2 &  16.9 &  18.1 \\
\bottomrule
\end{tabular}}
\end{table*}


\begin{table*}[t]
\centering
\caption{\textbf{Modality ablation} with Gemma~4 26B. Each row
is one input combination, marked by the checkmarks in the first four columns.
Throughout, \vcue{visual} is red (\vcue{Frm.}: frames, \vcue{Cap.}: visual
captions) and \acue{audio} is blue (\acue{Sub.}: subtitles, \acue{Aud.}: audio
analysis). Question groups are as in Table~\ref{tab:sota-judgement}. The winning score per question category across all ablations is put in bold.
\textbf{Single-pass graded scores (0--100).}}
\label{tab:ablations-judgement}
\scriptsize
\setlength{\tabcolsep}{2pt}
\resizebox{\textwidth}{!}{%
\begin{tabular}{cccc @{\hspace{2.5em}} cccc @{\hspace{2.5em}} cccc @{\hspace{1.5em}} cccc @{\hspace{1.5em}} cccc @{\hspace{2.5em}} c}
\toprule
\multicolumn{4}{c@{\hspace{2.5em}}}{\textbf{Inputs}}
      & \multicolumn{4}{c@{\hspace{2.5em}}}{\textbf{Simple}} & \multicolumn{13}{c}{\textbf{Specific}} \\
\cmidrule(l{0pt}r{2.5em}){1-4} \cmidrule(l{0pt}r{2.5em}){5-8} \cmidrule(l{0pt}r{0pt}){9-21}
  \vcue{Frm.} & \vcue{Cap.} & \acue{Sub.} & \acue{Aud.}
      & G & \vcue{V} & \acue{A} & Avg.
      & G1 & G2 & G3 & Avg.
      & \vcue{V1} & \vcue{V2} & \vcue{V3} & Avg.
      & \acue{A1} & \acue{A2} & \acue{A3} & Avg. & Avg. \\
\midrule
\vcue{\cmark{}} & \vcue{\xmark{}} & \acue{\xmark{}} & \acue{\xmark{}} &  25.3 &  36.3 &   0.2 &  20.6 &  29.5 &  26.4 &  16.5 &  24.2 &  25.3 &  21.9 &  19.1 &  22.1 &   0.4 &   0.4 &   7.5 &   2.7 &  16.3 \\
\vcue{\xmark{}} & \vcue{\cmark{}} & \acue{\xmark{}} & \acue{\xmark{}} &  22.0 &  31.9 &   0.3 &  18.1 &  22.8 &  22.4 &  21.0 &  22.1 &  25.6 &  31.0 &  20.3 &  25.6 &   0.6 &   0.1 &   5.1 &   1.9 &  16.6 \\
\vcue{\xmark{}} & \vcue{\xmark{}} & \acue{\cmark{}} & \acue{\xmark{}} &  41.4 &   1.2 &  20.2 &  20.9 &  13.4 &  47.1 &  34.5 &  31.7 &   0.4 &   0.0 &   0.3 &   0.2 &   5.8 &   3.1 &  43.3 &  17.4 &  16.4 \\
\vcue{\xmark{}} & \vcue{\xmark{}} & \acue{\xmark{}} & \acue{\cmark{}} &   3.1 &   0.0 &  38.1 &  13.7 &   9.1 &   0.9 &   1.6 &   3.9 &   0.0 &   0.0 &   0.0 &   0.0 &  46.1 &  14.9 &   0.9 &  20.6 &   8.2 \\
\addlinespace
\vcue{\cmark{}} & \vcue{\cmark{}} & \acue{\xmark{}} & \acue{\xmark{}} &  28.1 &  39.8 &   1.1 &  23.0 &  29.8 &  30.2 &  24.2 &  28.1 &  31.3 &  35.2 &  27.6 &  31.4 &   0.8 &   0.2 &   7.8 &   2.9 &  20.8 \\
\vcue{\cmark{}} & \vcue{\xmark{}} & \acue{\cmark{}} & \acue{\xmark{}} &  49.7 &  44.2 &  25.4 &  39.8 &  36.2 &  53.3 &  40.3 &  43.3 &  36.2 &  33.0 &  29.6 &  33.0 &   7.5 &   4.7 &  42.1 &  18.1 &  31.4 \\
\vcue{\cmark{}} & \vcue{\xmark{}} & \acue{\xmark{}} & \acue{\cmark{}} &   9.0 &  19.1 &  38.6 &  22.2 &  10.6 &  10.2 &  10.2 &  10.3 &   7.0 &   9.9 &   5.4 &   7.5 &  45.6 &  18.2 &   5.5 &  23.1 &  13.6 \\
\vcue{\xmark{}} & \vcue{\cmark{}} & \acue{\cmark{}} & \acue{\xmark{}} &  49.5 &  37.1 &  20.7 &  35.7 &  33.3 &  53.3 &  43.0 &  43.2 &  31.8 &  35.6 &  26.4 &  31.3 &   6.6 &   1.6 &  45.9 &  18.0 &  30.8 \\
\vcue{\xmark{}} & \vcue{\cmark{}} & \acue{\xmark{}} & \acue{\cmark{}} &  19.0 &  28.8 &  39.4 &  29.1 &  24.2 &  16.8 &  20.7 &  20.6 &  21.7 &  29.5 &  19.3 &  23.5 &  45.6 &  17.5 &   6.0 &  23.0 &  22.4 \\
\vcue{\xmark{}} & \vcue{\xmark{}} & \acue{\cmark{}} & \acue{\cmark{}} &  43.5 &   0.1 &  \textbf{51.2} &  31.6 &  17.7 &  47.4 &  35.6 &  33.6 &   0.1 &   0.0 &   0.1 &   0.0 &  \textbf{52.1} &  19.2 &  43.0 &  38.1 &  23.9 \\
\addlinespace
\vcue{\cmark{}} & \vcue{\cmark{}} & \acue{\cmark{}} & \acue{\xmark{}} &  50.6 &  \textbf{46.3} &  23.6 &  40.2 &  \textbf{36.7} &  \textbf{54.5} &  42.5 &  \textbf{44.6} &  \textbf{44.4} &  \textbf{41.2} &  \textbf{38.7} &  \textbf{41.4} &   7.3 &   1.8 &  42.8 &  17.3 &  34.4 \\
\vcue{\cmark{}} & \vcue{\cmark{}} & \acue{\xmark{}} & \acue{\cmark{}} &  18.7 &  32.8 &  39.4 &  30.3 &  23.2 &  19.5 &  19.6 &  20.7 &  25.2 &  32.4 &  18.5 &  25.4 &  45.1 &  17.5 &   8.6 &  23.7 &  23.3 \\
\vcue{\cmark{}} & \vcue{\xmark{}} & \acue{\cmark{}} & \acue{\cmark{}} &  44.8 &  37.9 &  44.7 &  42.4 &  23.1 &  50.5 &  39.5 &  37.7 &  34.3 &  32.6 &  26.8 &  31.2 &  50.0 &  19.3 &  41.0 &  36.7 &  35.2 \\
\vcue{\xmark{}} & \vcue{\cmark{}} & \acue{\cmark{}} & \acue{\cmark{}} &  \textbf{51.8} &  37.4 &  49.1 &  \textbf{46.1} &  34.0 &  53.2 &  \textbf{44.8} &  44.0 &  27.6 &  35.6 &  23.2 &  28.8 &  51.2 &  \textbf{19.6} &  \textbf{47.1} &  \textbf{39.3} &  37.4 \\
\addlinespace
\vcue{\cmark{}} & \vcue{\cmark{}} & \acue{\cmark{}} & \acue{\cmark{}} &  49.0 &  42.0 &  43.2 &  44.7 &  34.0 &  52.0 &  42.3 &  42.8 &  38.3 &  40.7 &  35.1 &  38.0 &  49.4 &  18.7 &  44.6 &  37.6 &  \textbf{39.5} \\
\bottomrule
\end{tabular}}
\end{table*}


\begin{table*}[htp!]
\centering
\caption{\textbf{Modality ablation} with Gemma~4 26B. Each row
is one input combination, marked by the checkmarks in the first four columns.
Throughout, \vcue{visual} is red (\vcue{Frm.}: frames, \vcue{Cap.}: visual
captions) and \acue{audio} is blue (\acue{Sub.}: subtitles, \acue{Aud.}: audio
analysis). Question groups are as in Table~\ref{tab:sota-judgement}. The winning score per question category across all ablations is put in bold.
\textbf{Checklist scores (0--100).}}
\label{tab:ablations-checklist}
\scriptsize
\setlength{\tabcolsep}{2pt}
\resizebox{\textwidth}{!}{%
\begin{tabular}{cccc @{\hspace{2.5em}} cccc @{\hspace{2.5em}} cccc @{\hspace{1.5em}} cccc @{\hspace{1.5em}} cccc @{\hspace{2.5em}} c}
\toprule
\multicolumn{4}{c@{\hspace{2.5em}}}{\textbf{Inputs}}
      & \multicolumn{4}{c@{\hspace{2.5em}}}{\textbf{Simple}} & \multicolumn{13}{c}{\textbf{Specific}} \\
\cmidrule(l{0pt}r{2.5em}){1-4} \cmidrule(l{0pt}r{2.5em}){5-8} \cmidrule(l{0pt}r{0pt}){9-21}
  \vcue{Frm.} & \vcue{Cap.} & \acue{Sub.} & \acue{Aud.}
      & G & \vcue{V} & \acue{A} & Avg.
      & G1 & G2 & G3 & Avg.
      & \vcue{V1} & \vcue{V2} & \vcue{V3} & Avg.
      & \acue{A1} & \acue{A2} & \acue{A3} & Avg. & Avg. \\
\midrule
\vcue{\cmark{}} & \vcue{\xmark{}} & \acue{\xmark{}} & \acue{\xmark{}} &   2.6 &   2.6 &   0.0 &   1.7 &   2.4 &   5.0 &   1.7 &   3.1 &   4.8 &   2.8 &   4.0 &   3.9 &   0.2 &   0.1 &   2.3 &   0.9 &   2.6 \\
\vcue{\xmark{}} & \vcue{\cmark{}} & \acue{\xmark{}} & \acue{\xmark{}} &   2.1 &   1.6 &   0.0 &   1.2 &   1.9 &   4.3 &   2.1 &   2.8 &   3.7 &   3.3 &   2.1 &   3.0 &   0.1 &   0.0 &   0.4 &   0.2 &   2.0 \\
\vcue{\xmark{}} & \vcue{\xmark{}} & \acue{\cmark{}} & \acue{\xmark{}} &  10.2 &   0.0 &   2.8 &   4.3 &   1.7 &  17.0 &   4.3 &   7.7 &   0.1 &   0.0 &   0.1 &   0.1 &   1.2 &   0.5 &   7.7 &   3.1 &   3.6 \\
\vcue{\xmark{}} & \vcue{\xmark{}} & \acue{\xmark{}} & \acue{\cmark{}} &   0.4 &   0.0 &   7.8 &   2.7 &   1.2 &   0.1 &   0.0 &   0.5 &   0.0 &   0.0 &   0.0 &   0.0 &  15.4 &   1.7 &   0.5 &   5.8 &   2.1  \\
\addlinespace
\vcue{\cmark{}} & \vcue{\cmark{}} & \acue{\xmark{}} & \acue{\xmark{}} &   2.8 &   2.8 &   0.1 &   2.0 &   2.2 &   5.9 &   3.2 &   3.8 &   4.4 &   4.6 &   5.0 &   4.7 &   0.7 &   0.0 &   1.9 &   0.8 &   3.1 \\
\vcue{\cmark{}} & \vcue{\xmark{}} & \acue{\cmark{}} & \acue{\xmark{}} &  \textbf{11.8} &   \textbf{4.5} &   2.6 &   6.3 &   4.0 &  17.5 &   4.9 &   8.8 &   \textbf{8.3} &   5.7 &   6.1 &   6.7 &   1.4 &   0.7 &   8.5 &   3.5 &   6.4 \\
\vcue{\cmark{}} & \vcue{\xmark{}} & \acue{\xmark{}} & \acue{\cmark{}} &   0.6 &   0.8 &   7.3 &   2.9 &   1.1 &   1.7 &   0.7 &   1.2 &   1.1 &   1.1 &   0.7 &   1.0 &  13.0 &   1.7 &   1.7 &   5.4 &   2.5 \\
\vcue{\xmark{}} & \vcue{\cmark{}} & \acue{\cmark{}} & \acue{\xmark{}} &  11.4 &   3.3 &   3.0 &   5.9 &   \textbf{4.2} &  \textbf{19.8} &   \textbf{6.6} &   \textbf{10.2} &   6.2 &   4.3 &   4.0 &   4.8 &   1.3 &   0.3 &   8.3 &   3.3 &   6.1 \\
\vcue{\xmark{}} & \vcue{\cmark{}} & \acue{\xmark{}} & \acue{\cmark{}} &   2.1 &   1.0 &  10.6 &   4.5 &   2.2 &   2.9 &   1.9 &   2.3 &   3.4 &   2.4 &   2.2 &   2.7 &  16.1 &   1.9 &   1.1 &   6.4 &   3.8 \\
\vcue{\xmark{}} & \vcue{\xmark{}} & \acue{\cmark{}} & \acue{\cmark{}} &  10.0 &   0.0 &  11.4 &   7.1 &   2.1 &  16.5 &   4.4 &   7.7 &   0.0 &   0.0 &   0.0 &   0.0 &  16.5 &   2.1 &   7.9 &   8.8 &   5.5 \\
\addlinespace
\vcue{\cmark{}} & \vcue{\cmark{}} & \acue{\cmark{}} & \acue{\xmark{}} &  10.9 &   \textbf{4.5} &   2.5 &   6.0 &   3.5 &  18.3 &   5.4 &   9.0 &   8.2 &   6.2 &   \textbf{7.7} &   \textbf{7.4} &   1.3 &   0.3 &   8.2 &   3.2 &   6.5 \\
\vcue{\cmark{}} & \vcue{\cmark{}} & \acue{\xmark{}} & \acue{\cmark{}} &   1.8 &   2.0 &   9.4 &   4.4 &   1.9 &   3.7 &   1.4 &   2.4 &   3.6 &   3.1 &   3.0 &   3.2 &  14.2 &   1.7 &   2.2 &   6.0 &   3.9 \\
\vcue{\cmark{}} & \vcue{\xmark{}} & \acue{\cmark{}} & \acue{\cmark{}} &   9.9 &   3.3 &   9.0 &   7.4 &   2.4 &  17.6 &   4.7 &   8.2 &   6.0 &   6.0 &   4.8 &   5.6 &  14.7 &   \textbf{2.3} &   7.6 &   8.2 &   7.3 \\
\vcue{\xmark{}} & \vcue{\cmark{}} & \acue{\cmark{}} & \acue{\cmark{}} &  11.6 &   2.7 &   \textbf{12.3} &   \textbf{8.8} &   3.9 &  18.1 &   6.3 &   9.4 &   4.6 &   4.6 &   3.4 &   4.2 &   \textbf{16.6} &   2.1 &   \textbf{10.1} &   \textbf{9.6} &   7.7 \\
\addlinespace
\vcue{\cmark{}} & \vcue{\cmark{}} & \acue{\cmark{}} & \acue{\cmark{}} &  10.7 &   4.0 &  11.0 &   8.6 &   3.4 &  17.5 &   5.9 &   8.9 &   5.7 &   \textbf{6.8} &   5.8 &   6.1 &  14.5 &   1.8 &   9.2 &   8.5 &   \textbf{7.8} \\
\bottomrule
\end{tabular}}
\end{table*}

Tables~\ref{tab:sota-judgement} and~\ref{tab:sota-checklist}
report each model's performance under full context, i.e.\ all the available input channels: visual captions, subtitles, audio and keyframes.

The proprietary model Gemini-3.5-Flash performs the best overall ($62.0$ simple / $58.1$ specific) followed by Qwen3.5-27B ($46.4$ / $45.9$) and
Gemma~4~26B ($44.7$ / $39.5$). This ordering is largely consistent across
both metrics: the same three models occupy the top three positions under
checklist scoring (Table~\ref{tab:sota-checklist}), and InternVL3.5-4B is
the weakest model under both metrics.

\begin{figure}[t]
    \centering
    \begin{subfigure}[t]{\linewidth}
        \centering
        \includegraphics[width=\linewidth]{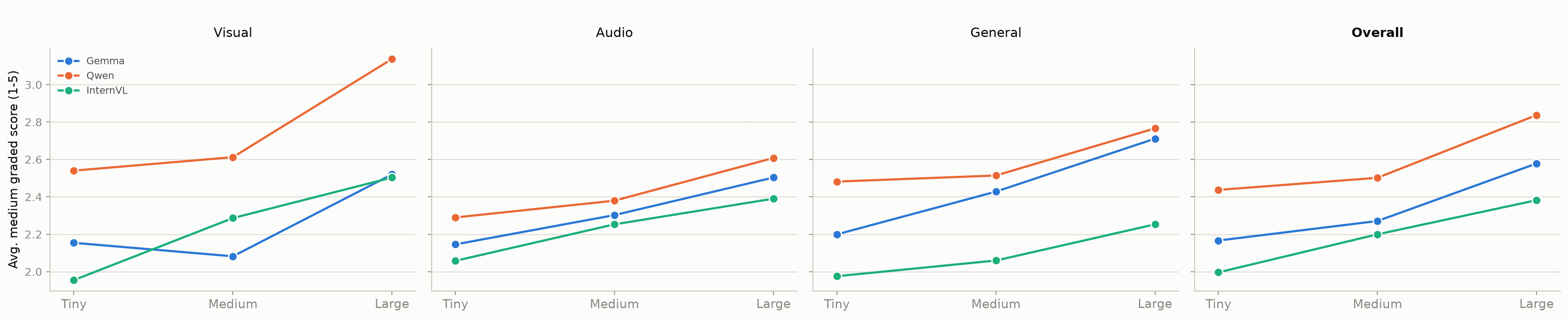}
        \caption{Single-pass graded score.}
        \label{fig:progression-judge}
    \end{subfigure}

    \vspace{0.3em}
    \begin{subfigure}[t]{\linewidth}
        \centering
        \includegraphics[width=\linewidth]{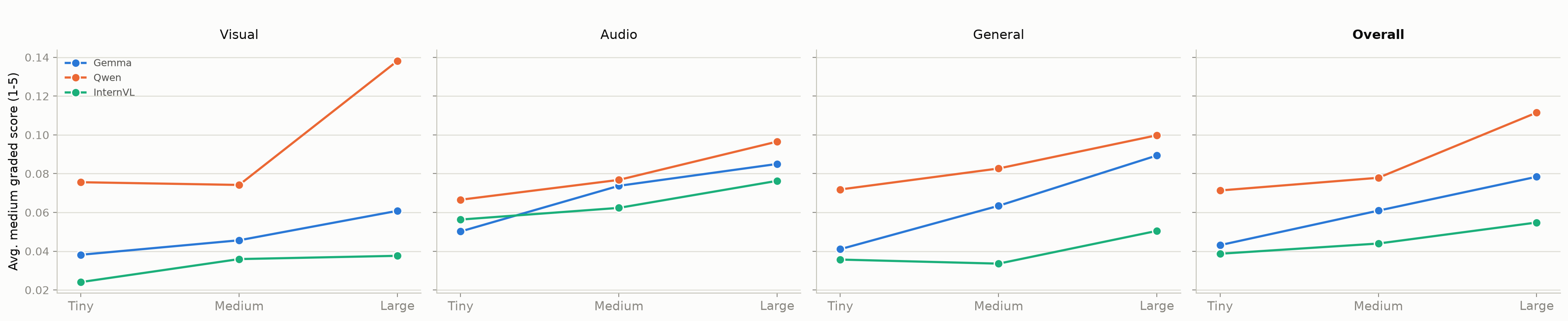}
        \caption{Checklist score.}
        \label{fig:progression-checklist}
    \end{subfigure}
    \caption{\textbf{Single-pass graded progression with model scale, per category.} Open-weight
    families under full context. 
    We see that scale generally helps, with few
    exceptions: Gemma dips on visual at 12B (\subref{fig:progression-judge}), and InternVL drops on visual and general at 8B and 30B (\subref{fig:progression-checklist}).}
    \label{fig:progression}
\end{figure}

Figure~\ref{fig:progression} demonstrates that, with limited exceptions (InternVL3.5-8B exceeds Gemma-4-12B in the single-pass graded score for visual questions), this pattern holds consistently across question category, meaning no model demonstrates a particular strength or weakness at any given category. 

\subsection{Contribution of individual modalities}
\label{sec:modality-ablation}

Tables~\ref{tab:ablations-judgement} and~\ref{tab:ablations-checklist}
report performance across all evaluated input combinations for
Gemma~4~26B, highlighting contributions of each modality combination.

\paragraph{Single modalities specialize almost perfectly, and fail almost completely outside their specialty.} Captions-only and frames-only score 0.3 and 0.2 on audio questions (graded), and audio-only scores 0.0 on visual questions; checklist judgement shows the same pattern (0.0 across all three). This confirms that \texttt{TAKE\,85}'s categories isolate modality-specific information, rather than being answerable from any single source or prior knowledge.

\paragraph{Audio questions benefit from text-based context.} Pairing audio with a second modality affects audio-question performance very differently depending on which is added for both single-pass and checklist: from audio-only (38.1 / 7.8), subtitles give the largest gain (51.2 / 11.4), captions a modest one (39.4 / 10.6), and frames essentially none (38.6 / 7.3).

\paragraph{The richest ablation does not perform best within specific domains.} Under single-pass scoring (Table~\ref{tab:ablations-judgement}), frames+captions+subtitles leads every visual column and two of three general columns, outperforming the full four-modality combination in each case; adding audio \emph{reduces} all five scores, confounding rather than helping. The audio columns show the reverse: no single combination wins all three, with captions+subtitles+audio leading A2/A3 and subtitles+audio leading A1.

These results support a central claim of this work: \textbf{directorial-intent understanding is not reducible to any single modality}. Audio questions benefit substantially from textual evidence, while audio-only and visual-only ablations collapse almost completely on the opposite category. Each additional modality contributes evidence, but not every combination is equally useful for a given question.
\begin{figure}[htp!]
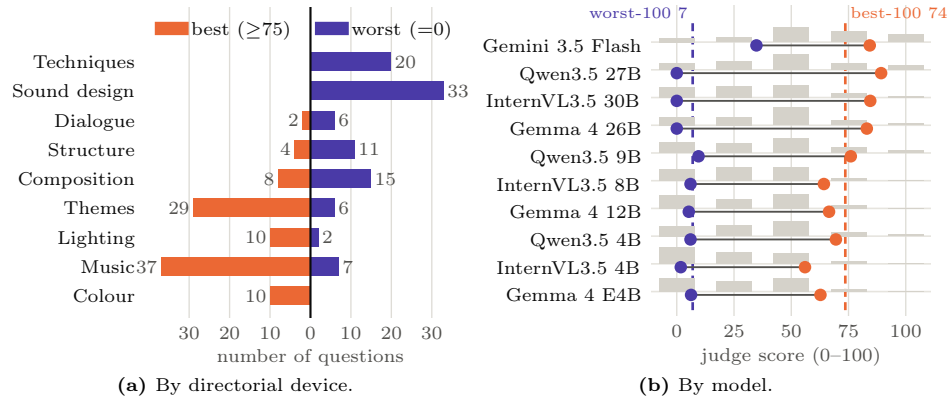

    \centering
    \begin{subfigure}[t]{0.49\linewidth}
        \centering
        \input{figs/bvw-devices.pgf}
        \caption{By directorial device.}
        \label{fig:bvw-devices}
    \end{subfigure}\hfill
    \begin{subfigure}[t]{0.49\linewidth}
        \centering
        \input{figs/bvw-models.pgf}
        \caption{By model.}
        \label{fig:bvw-models}
    \end{subfigure}
    \caption{\textbf{Best versus worst answered questions.} The 100 questions all
    three large models answer well (graded score $\geq$75) against the 100 they
    all fail (0), split by directorial device (\subref{fig:bvw-devices}) and
    positioned within each model's own score distribution
    (\subref{fig:bvw-models}, grey). The three large models define the sets, so
    their worst-set mean is 0 by construction.}
    \label{fig:quali-best-vs-worst}
\end{figure}

\subsection{Best versus worst answered questions}
\label{sec:quali}

As shown in Figure~\ref{fig:quali-best-vs-worst}, we select the questions on which the three large models agree, and contrast the 100 they all answer well (graded score $\geq$75) with the 100 they all fail (0). Two patterns emerge.

\paragraph{Difficulty follows the directorial device, not the modality.}
The two sets separate almost completely by device. All 33 sound-design questions and all 20 technique questions fall in the worst set, as do the majority of dialogue, structure and composition questions. Color, music, lighting and themes questions are answered well in over 80\% of cases. The intermediate range is essentially empty. This ordering is independent of our category labels: sound design and music are both audio questions answered from the same channel yet occupy opposite extremes, as do composition and color. Competence is therefore organized by directorial device rather than by input modality.

\paragraph{The worst set is not intrinsically unanswerable.}
Because the three large models define the sets, the informative comparison is with the remaining models. The six open-weight models that did not define the worst set still collapse on it, averaging below 10 and answering at most 10 of the 100 questions acceptably. Gemini~3.5~Flash does not: of the 97 questions for which it returned a score, it averages 35 and answers 41 acceptably. The proprietary advantage is concentrated precisely where open-weight models almost fail completely, indicating a shared blind spot rather than a limit of the questions themselves.

\section{Conclusion}
\label{sec:conclusion}
We introduced \textbf{\texttt{TAKE\,85}}, the first benchmark for evaluating directorial intent in films. Our experiments show that, despite impressive progress in perceptual video understanding, current MLLMs struggle to reason about the creative decisions that shape a viewer's interpretation. The failure is systematic: no input modality is sufficient on its own, and difficulty follows the directorial device rather than the modality, with sound design and directorial technique defeating every open-weight model we evaluate. This reveals a fundamental gap between recognizing audiovisual content and understanding its intended meaning. We hope \texttt{TAKE\,85} encourages the development of multimodal models that reason not only about \emph{what} is shown, but also about \emph{why} it is presented that way.

\newpage
\section{Acknowledgement}
This work was supported by Hi! Paris (grant and fellowship), the ANR/France 2030 program (ANR-23-IACL-0005), the ANR JCJC projects "The Why behind scenes" (ANR-22-CE23-0007), ANR JCJC "REEL-WORLD", and a Google DeepMind academic gift. Computing resources were provided by GENCI through access to the IDRIS HPC facilities under allocation 2026-AD011014300R3, and by Google Gemini.

\bibliographystyle{splncs04}
\bibliography{main}

@String(IJCV  = {Int. J. Comput. Vis.})

@String(CVPR  = {IEEE Conf. Comput. Vis. Pattern Recog.})

@String(ICCV  = {Int. Conf. Comput. Vis.})

@String(ECCV  = {Eur. Conf. Comput. Vis.})

@String(NeurIPS = {Adv. Neural Inform. Process. Syst.})

@String(ACCV  = {Asian Conf. Comput. Vis.})

@String(AAAI  = {AAAI})

@String(IJCV  = {IJCV})

@String(CVPR  = {CVPR})

@String(ICCV  = {ICCV})

@String(ICCVW  = {ICCVW})

@String(ECCV  = {ECCV})

@String(NeurIPS = {NeurIPS})

@String(ACCV  = {ACCV})

@String(EMNLP    = {EMNLP})

@article{ghermi2024long,
  title   = {Long story short: Story-level video understanding from 20k short films},
  author  = {Ghermi, Ridouane and Wang, Xi and Kalogeiton, Vicky and Laptev, Ivan},
  journal = {arXiv preprint arXiv:2406.10221},
  year    = {2024}
}

@inproceedings{xie2025shot,
  title     = {Shot-by-Shot: Film-Grammar-Aware Training-Free Audio Description Generation},
  author    = {Xie, Junyu and Han, Tengda and Bain, Max and Nagrani, Arsha and Khandelwal, Eshika and Varol, G{\"u}l and Xie, Weidi and Zisserman, Andrew},
  booktitle = {Proceedings of the IEEE/CVF International Conference on Computer Vision},
  pages     = {16503--16513},
  year      = {2025}
}

@article{tang2025video,
  title   = {Video understanding with large language models: A survey},
  author  = {Tang, Yunlong and Bi, Jing and Xu, Siting and Song, Luchuan and Liang, Susan and Wang, Teng and Zhang, Daoan and An, Jie and Lin, Jingyang and Zhu, Rongyi and others},
  journal = {Transactions on Circuits and Systems for Video Technology},
  year    = {2025}
}

@article{kumar2025videollm,
  title   = {VideoLLM Benchmarks and Evaluation: A Survey},
  author  = {Kumar, Yogesh},
  journal = {arXiv preprint arXiv:2505.03829},
  year    = {2025}
}

@inproceedings{krishna2017dense,
  title     = {Dense-captioning events in videos},
  author    = {Krishna, Ranjay and Hata, Kenji and Ren, Frederic and Fei-Fei, Li and Carlos Niebles, Juan},
  booktitle = iccv,
  year      = {2017}
}

@inproceedings{zhou2018end,
  title     = {End-to-end dense video captioning with masked transformer},
  author    = {Zhou, Luowei and Zhou, Yingbo and Corso, Jason J and Socher, Richard and Xiong, Caiming},
  booktitle = cvpr,
  year      = {2018}
}

@inproceedings{wang2018bidirectional,
  title     = {Bidirectional attentive fusion with context gating for dense video captioning},
  author    = {Wang, Jingwen and Jiang, Wenhao and Ma, Lin and Liu, Wei and Xu, Yong},
  booktitle = cvpr,
  year      = {2018}
}

@inproceedings{yang2023vid2seq,
  title     = {Vid2seq: Large-scale pretraining of a visual language model for dense video captioning},
  author    = {Yang, Antoine and Nagrani, Arsha and Seo, Paul Hongsuck and Miech, Antoine and Pont-Tuset, Jordi and Laptev, Ivan and Sivic, Josef and Schmid, Cordelia},
  booktitle = cvpr,
  year      = {2023}
}

@inproceedings{islam2024video,
  title     = {Video recap: Recursive captioning of hour-long videos},
  author    = {Islam, Md Mohaiminul and Ho, Ngan and Yang, Xitong and Nagarajan, Tushar and Torresani, Lorenzo and Bertasius, Gedas},
  booktitle = cvpr,
  year      = {2024}
}

@inproceedings{zhou2024streaming,
  title     = {Streaming dense video captioning},
  author    = {Zhou, Xingyi and Arnab, Anurag and Buch, Shyamal and Yan, Shen and Myers, Austin and Xiong, Xuehan and Nagrani, Arsha and Schmid, Cordelia},
  booktitle = cvpr,
  year      = {2024}
}

@inproceedings{han2023autoad,
  title     = {Autoad: Movie description in context},
  author    = {Han, Tengda and Bain, Max and Nagrani, Arsha and Varol, G{\"u}l and Xie, Weidi and Zisserman, Andrew},
  booktitle = cvpr,
  year      = {2023}
}

@inproceedings{han2023autoadii,
  title     = {Autoad ii: The sequel-who, when, and what in movie audio description},
  author    = {Han, Tengda and Bain, Max and Nagrani, Arsha and Varol, Gul and Xie, Weidi and Zisserman, Andrew},
  booktitle = iccv,
  year      = {2023}
}

@inproceedings{han2024autoad,
  title     = {Autoad iii: The prequel-back to the pixels},
  author    = {Han, Tengda and Bain, Max and Nagrani, Arsha and Varol, G{\"u}l and Xie, Weidi and Zisserman, Andrew},
  booktitle = cvpr,
  year      = {2024}
}

@inproceedings{xie2024autoad,
  title     = {Autoad-zero: A training-free framework for zero-shot audio description},
  author    = {Xie, Junyu and Han, Tengda and Bain, Max and Nagrani, Arsha and Varol, G{\"u}l and Xie, Weidi and Zisserman, Andrew},
  booktitle = accv,
  year      = {2024}
}

@inproceedings{fang2025distinctad,
  title     = {DistinctAD: Distinctive audio description generation in contexts},
  author    = {Fang, Bo and Wu, Wenhao and Wu, Qiangqiang and Song, Yuxin and Chan, Antoni B},
  booktitle = cvpr,
  year      = {2025}
}

@inproceedings{wang2025contextual,
  title     = {Contextual ad narration with interleaved multimodal sequence},
  author    = {Wang, Hanlin and Tong, Zhan and Zheng, Kecheng and Shen, Yujun and Wang, Limin},
  booktitle = cvpr,
  year      = {2025}
}

@inproceedings{zhang2024mm,
  title     = {Mm-narrator: Narrating long-form videos with multimodal in-context learning},
  author    = {Zhang, Chaoyi and Lin, Kevin and Yang, Zhengyuan and Wang, Jianfeng and Li, Linjie and Lin, Chung-Ching and Liu, Zicheng and Wang, Lijuan},
  booktitle = cvpr,
  year      = {2024}
}

@article{wang2024qwen2,
  title   = {Qwen2-vl: Enhancing vision-language model's perception of the world at any resolution},
  author  = {Wang, Peng and Bai, Shuai and Tan, Sinan and Wang, Shijie and Fan, Zhihao and Bai, Jinze and Chen, Keqin and Liu, Xuejing and Wang, Jialin and Ge, Wenbin and others},
  journal = {arXiv preprint arXiv:2409.12191},
  year    = {2024}
}

@article{bai2025qwen3,
  title   = {Qwen3-vl technical report},
  author  = {Bai, Shuai and Cai, Yuxuan and Chen, Ruizhe and Chen, Keqin and Chen, Xionghui and Cheng, Zesen and Deng, Lianghao and Ding, Wei and Gao, Chang and Ge, Chunjiang and others},
  journal = {arXiv preprint arXiv:2511.21631},
  year    = {2025}
}

@misc{qwen3.5,
    title  = {{Qwen3.5}: Towards Native Multimodal Agents},
    author = {{Qwen Team}},
    month  = {February},
    year   = {2026},
    url    = {https://qwen.ai/blog?id=qwen3.5}
}

@article{wang2025internvl3,
  title   = {InternVL3.5: Advancing open-source multimodal models in versatility, reasoning, and efficiency},
  author  = {Wang, Weiyun and Gao, Zhangwei and Gu, Lixin and Pu, Hengjun and Cui, Long and Wei, Xingguang and Liu, Zhaoyang and Jing, Linglin and Ye, Shenglong and Shao, Jie and others},
  journal = {arXiv preprint arXiv:2508.18265},
  year    = {2025}
}

@article{zhang2025videollama,
  title   = {Videollama 3: Frontier multimodal foundation models for image and video understanding},
  author  = {Zhang, Boqiang and Li, Kehan and Cheng, Zesen and Hu, Zhiqiang and Yuan, Yuqian and Chen, Guanzheng and Leng, Sicong and Jiang, Yuming and Zhang, Hang and Li, Xin and others},
  journal = {arXiv preprint arXiv:2501.13106},
  year    = {2025}
}

@article{achiam2023gpt,
  title   = {Gpt-4 technical report},
  author  = {Achiam, Josh and Adler, Steven and Agarwal, Sandhini and Ahmad, Lama and Akkaya, Ilge and Aleman, Florencia Leoni and Almeida, Diogo and Altenschmidt, Janko and Altman, Sam and Anadkat, Shyamal and others},
  journal = {arXiv preprint arXiv:2303.08774},
  year    = {2023}
}

@article{mao2026cinecap,
  title   = {CineCap: Structured Reasoning with Spatio-Temporal Anchors for Cinematographic Video Captioning},
  author  = {Mao, Xinyu and Zeng, Yuhui and Liu, Xiaokun and Qin, Wenyu and Wang, Meng and Tao, Xin and Wan, Pengfei and Xing, Xiaohan and Meng, Max},
  journal = {arXiv preprint arXiv:2606.24636},
  year    = {2026}
}

@inproceedings{tang2025vidcomposition,
  title     = {Vidcomposition: Can mllms analyze compositions in compiled videos?},
  author    = {Tang, Yunlong and Guo, Junjia and Hua, Hang and Liang, Susan and Feng, Mingqian and Li, Xinyang and Mao, Rui and Huang, Chao and Bi, Jing and Zhang, Zeliang and others},
  booktitle = cvpr,
  year      = {2025}
}

@inproceedings{tapaswi2016movieqa,
  title     = {Movieqa: Understanding stories in movies through question-answering},
  author    = {Tapaswi, Makarand and Zhu, Yukun and Stiefelhagen, Rainer and Torralba, Antonio and Urtasun, Raquel and Fidler, Sanja},
  booktitle = iccv,
  year      = {2016}
}

@inproceedings{lei2018tvqa,
  title     = {Tvqa: Localized, compositional video question answering},
  author    = {Lei, Jie and Yu, Licheng and Bansal, Mohit and Berg, Tamara},
  booktitle = emnlp,
  year      = {2018}
}

@inproceedings{yu2019activitynet,
  title     = {Activitynet-qa: A dataset for understanding complex web videos via question answering},
  author    = {Yu, Zhou and Xu, Dejing and Yu, Jun and Yu, Ting and Zhao, Zhou and Zhuang, Yueting and Tao, Dacheng},
  booktitle = aaai,
  year      = {2019}
}

@inproceedings{xu2017video,
  title     = {Video Question Answering via Gradually Refined Attention over Appearance and Motion},
  author    = {Xu, Dejing and Zhao, Zhou and Xiao, Jun and Wu, Fei and Zhang, Hanwang and He, Xiangnan and Zhuang, Yueting},
  booktitle = {ACM Multimedia},
  year      = {2017}
}

@inproceedings{li2022representation,
  title     = {From representation to reasoning: Towards both evidence and commonsense reasoning for video question-answering},
  author    = {Li, Jiangtong and Niu, Li and Zhang, Liqing},
  booktitle = cvpr,
  year      = {2022}
}

@inproceedings{a2026cineaste,
  title     = {Cineaste: A Fine-grained Contextual Movie Question Answering Benchmark with Automated Data Curation},
  author    = {A Shah, Nisarg and Ziai, Amir and Ekanadham, Chaitanya and Patel, Vishal M},
  booktitle = cvpr,
  year      = {2026}
}

@inproceedings{wu2021towards,
  title     = {Towards long-form video understanding},
  author    = {Wu, Chao-Yuan and Krahenbuhl, Philipp},
  booktitle = cvpr,
  year      = {2021}
}

@article{rawal2024cinepile,
  title   = {Cinepile: A long video question answering dataset and benchmark},
  author  = {Rawal, Ruchit and Saifullah, Khalid and Farr{\'e}, Miquel and Basri, Ronen and Jacobs, David and Somepalli, Gowthami and Goldstein, Tom},
  journal = {arXiv preprint arXiv:2405.08813},
  year    = {2024}
}

@article{wang2026cinetechbench,
  title   = {Cinetechbench: A benchmark for cinematographic technique understanding and generation},
  author  = {Wang, Xinran and Xu, Songyu and Xiangxuan, Shan and Zhang, Yuxuan and Diao, Muxi and Duan, Xueyan and Liang, Kongming and Ma, Zhanyu and others},
  journal = neurips,
  year    = {2026}
}

@article{liu2026shotbench,
  title   = {Shotbench: Expert-level cinematic understanding in vision-language models},
  author  = {Liu, Hongbo and He, Jingwen and Jin, Yi and Zheng, Dian and Dong, Yuhao and Zhang, Fan and Huang, Ziqi and He, Yinan and Chen, Weichao and Qiao, Yu and others},
  journal = neurips,
  year    = {2026}
}

@article{lin2026towards,
  title   = {Towards understanding camera motions in any video},
  author  = {Lin, Zhiqiu and Cen, Siyuan and Jiang, Daniel and Karhade, Jay and Wang, Hewei and Mitra, Chancharik and Ling, Yu Tong Tiffany and Huang, Yuhan and Zawar, Rushikesh and Bai, Xue and others},
  journal = neurips,
  year    = {2026}
}

@inproceedings{courant2021high,
  title     = {High-level features for movie style understanding},
  author    = {Courant, Robin and Lino, Christophe and Christie, Marc and Kalogeiton, Vicky},
  booktitle = iccvw,
  year      = {2021}
}

@article{liu2024funnynet,
  title   = {Funnynet-w: Multimodal learning of funny moments in videos in the wild},
  author  = {Liu, Zhi-Song and Courant, Robin and Kalogeiton, Vicky},
  journal = ijcv,
  year    = {2024}
}

@inproceedings{liu2022funnynet,
  title     = {Funnynet: Audiovisual learning of funny moments in videos},
  author    = {Liu, Zhisong and Courant, Robin and Kalogeiton, Vicky},
  booktitle = accv,
  year      = {2022}
}

@article{canini2013classifying,
  title     = {Classifying cinematographic shot types},
  author    = {Canini, Luca and Benini, Sergio and Leonardi, Riccardo},
  journal   = {Multimedia tools and applications},
  year      = {2013},
  publisher = {Springer}
}

@inproceedings{rao2020unified,
  title     = {A Unified Framework for Shot Type Classification Based on Subject Centric Lens},
  author    = {Rao, Anyi and Wang, Jiaze and Xu, Linning and Jiang, Xuekun and Huang, Qingqiu and Zhou, Bolei and Lin, Dahua},
  booktitle = eccv,
  year      = {2020}
}

@inproceedings{huang2020movienet,
  title     = {MovieNet: A Holistic Dataset for Movie Understanding},
  author    = {Huang, Qingqiu and Xiong, Yu and Rao, Anyi and Wang, Jiaze and Lin, Dahua},
  booktitle = eccv,
  year      = {2020}
}

@article{huang2018trailers,
  title   = {From Trailers to Storylines: An Efficient Way to Learn from Movies},
  author  = {Huang, Qingqiu and Xiong, Yuanjun and Xiong, Yu and Zhang, Yuqi and Lin, Dahua},
  journal = {arXiv preprint arXiv:1806.05341},
  year    = {2018}
}

@inproceedings{savardi2023recognition,
  title     = {Recognition of camera angle and camera level in movies from single frames},
  author    = {Savardi, Mattia and Kov{\'a}cs, Andr{\'a}s B{\'a}lint and Signoroni, Alberto and Benini, Sergio},
  booktitle = {ACM International Conference on Interactive Media Experiences Workshops},
  year      = {2023}
}

@article{jiang2021jointly,
  title   = {Jointly learning the attributes and composition of shots for boundary detection in videos},
  author  = {Jiang, Xuekun and Jin, Libiao and Rao, Anyi and Xu, Linning and Lin, Dahua},
  journal = {Transactions on Multimedia},
  year    = {2021}
}

@article{li2023toward,
  title   = {Toward unified and quantitative cinematic shot attribute analysis},
  author  = {Li, Yuzhi and Tian, Feng and Xu, Haojun and Lu, Tianfeng},
  journal = {Electronics},
  year    = {2023}
}

@article{lu2024exploring,
  title   = {Exploring challenge and explainable shot type classification using SAM-guided approaches},
  author  = {Lu, Fengtian and Li, Yuzhi and Tian, Feng},
  journal = {Signal, Image and Video Processing},
  year    = {2024}
}

@article{vacchetti2022cinematographic,
  title   = {Cinematographic shot classification with deep ensemble learning},
  author  = {Vacchetti, Bartolomeo and Cerquitelli, Tania},
  journal = {Electronics},
  year    = {2022}
}

@article{barriere2025standup4ai,
  title   = {StandUp4AI: A new multilingual dataset for humor detection in stand-up comedy videos},
  author  = {Barriere, Valentin and Gomez, Nahuel and Hemamou, Leo and Callejas, Sofia and Ravenet, Brian},
  journal = {arXiv preprint arXiv:2505.18903},
  volume  = {2},
  year    = {2025}
}

@inproceedings{hanania2026mtllfm,
  title     = {MTLLFM: Multimodal-Temporal Laughter Localization: UR-FUNNY-Temporal and SMILE-Temporal Benchmarks with an Adaptive Multimodal Fusion Model},
  author    = {Hanania, Eyal and Kirsch, Nadav and Arkushin, Daniel and Benvenisti, Jonathan and Bercovich, Amos and Zemmour, Elie and Froim, Sahar},
  booktitle = cvpr,
  year      = {2026}
}

@misc{omeleto_youtube,
  author       = {{Omeleto}},
  title        = {Omeleto},
  howpublished = {\url{https://www.youtube.com/@Omeleto}},
  note         = {YouTube}
}

@article{goel2025audio,
  title={Audio flamingo 3: Advancing audio intelligence with fully open large audio language models},
  author={Goel, Arushi and Ghosh, Sreyan and Kim, Jaehyeon and Kumar, Sonal and Kong, Zhifeng and Lee, Sang-gil and Yang, Chao-Han Huck and Duraiswami, Ramani and Manocha, Dinesh and Valle, Rafael and others},
  journal={arXiv preprint arXiv:2507.08128},
  year={2025}
}

@article{shi2025sam,
  title={Sam audio: Segment anything in audio},
  author={Shi, Bowen and Tjandra, Andros and Hoffman, John and Wang, Helin and Wu, Yi-Chiao and Gao, Luya and Richter, Julius and Le, Matt and Vyas, Apoorv and Chen, Sanyuan and others},
  journal={arXiv preprint arXiv:2512.18099},
  year={2025}
}

@inproceedings{radford2023robust,
  title={Robust speech recognition via large-scale weak supervision},
  author={Radford, Alec and Kim, Jong Wook and Xu, Tao and Brockman, Greg and McLeavey, Christine and Sutskever, Ilya},
  booktitle={International conference on machine learning},
  pages={28492--28518},
  year={2023},
  organization={PMLR}
}

@misc{claude_sonnet,
  author       = {{Anthropic}},
  title        = {Claude Sonnet 4.6},
  year         = {2026},
  howpublished = {\url{https://www.anthropic.com/claude}},
  note         = {Large language model. Accessed: 2026-07-29}
}

@article{team2026gemma,
  title={Gemma 4 technical report},
  author={Team, Gemma and Abd, Sherif El and Aggarwal, Vaibhav and Algayres, Robin and Andreev, Alek and Bachem, Olivier and Ballantyne, Ian and Brick, Cormac and C{\u{a}}rbune, Victor and Casbon, Michelle and others},
  journal={arXiv preprint arXiv:2607.02770},
  year={2026}
}

@misc{google_gemini_3_5_flash,
  author       = {{Google DeepMind}},
  title        = {Gemini 3.5 Flash},
  year         = {2026},
  howpublished = {\url{https://deepmind.google/models/model-cards/gemini-3-5-flash/}},
}

@article{meta2024llama,
  title={Llama 3.2: Revolutionizing edge ai and vision with open, customizable models},
  author={{Meta AI}},
  journal={Meta AI Blog},
  year={2024}
}
\end{document}